\documentclass[10pt, conference, compsocconf]{IEEEtran}
\IEEEoverridecommandlockouts
\usepackage{common}
\def\BibTeX{{\rm B\kern-.05em{\sc i\kern-.025em b}\kern-.08em
    T\kern-.1667em\lower.7ex\hbox{E}\kern-.125emX}}
\begin{document}

\title{DaiMoN: A Decentralized Artificial Intelligence Model Network\\
}
\author{
\IEEEauthorblockN{Surat Teerapittayanon}
\IEEEauthorblockA{
\textit{Harvard University}\\
Cambridge, USA\\
steerapi@seas.harvard.edu}
\and
\IEEEauthorblockN{H. T. Kung}
\IEEEauthorblockA{
\textit{Harvard University}\\
Cambridge, USA\\
kung@harvard.edu}
}

\maketitle

\begin{abstract}
We introduce DaiMoN, a decentralized artificial intelligence model network, which incentivizes peer collaboration in improving the accuracy of machine learning models for a given classification problem. It is an autonomous network where peers may submit models with improved accuracy and other peers may verify the accuracy improvement. The system maintains an append-only decentralized ledger to keep the log of critical information, including who has trained the model and improved its accuracy, when it has been improved, by how much it has improved, and where to find the newly updated model. DaiMoN rewards these contributing peers with cryptographic tokens. A main feature of DaiMoN is that it allows peers to verify the accuracy improvement of submitted models without knowing the test labels. This is an essential component in order to mitigate intentional model overfitting by model-improving peers. To enable this model accuracy evaluation with hidden test labels, DaiMoN uses a novel learnable \textit{Distance Embedding for Labels} (DEL) function proposed in this paper. Specific to each test dataset, DEL scrambles the test label vector by embedding it in a low-dimension space while approximately preserving the distance between the dataset's test label vector and a label vector inferred by the classifier. It therefore allows \textit{proof-of-improvement} (PoI) by peers without providing them access to true test labels. We provide analysis and empirical evidence that under DEL, peers can accurately assess model accuracy. We also argue that it is hard to invert the embedding function and thus, DEL is resilient against attacks aiming to recover test labels in order to cheat. Our prototype implementation of DaiMoN is available at \url{https://github.com/steerapi/daimon}.
\end{abstract}


\section{Introduction}
Network-based services are at the intersection of a revolution. Many centralized monolithic services are being replaced with decentralized microservices. The utility of decentralized ledgers showcases this change, and has been demonstrated by the usage of Bitcoin~\cite{nakamoto2008bitcoin} and Ethereum~\cite{wood2014ethereum}.

The same trend towards decentralization is expected to affect the field of artificial intelligence (AI), and in particular machine learning, as well. Complex models such as deep neural networks require large amounts of computational power and resources to train. Yet, these large, complex models are being retrained over and over again by different parties for similar performance objectives, wasting computational power and resources. Currently, only a relatively small number of pretrained models such as pretrained VGG~\cite{simonyan2014very}, ResNet~\cite{he2016deep}, GoogLeNet~\cite{szegedy2015going}, and BERT~\cite{devlin2018bert} are made available for reuse. 

One reason for this is that the current system to share models is centralized, limiting both the number of available models and incentives for people to participate and share models. Examples of these centralized types of systems are Caffe Model Zoo~\cite{bvlcmodelzoo}, Pytorch Model Zoo~\cite{pytorchmodelzoo}, Tensorflow Model Zoo\cite{tensorflowzoo}, and modelzoo.co~\cite{modelzooco}.

In other fields seeking to incentivize community participation, cryptocurrencies and cryptographic tokens based on decentralized ledger technology (DLT) have been used~\cite{nakamoto2008bitcoin, wood2014ethereum}. In addition to incentives, DLT offers the potential to support transparency, traceability, and digital trust at scale. The ledger is append-only, immutable, public, and can be audited and validated by anyone without a trusted third-party.

In this paper, we introduce DaiMoN, a decentralized artificial intelligence model network that brings the benefits of DLT to the field of machine learning. DaiMoN uses DLT and a token-based economy to incentivize people to improve machine learning models. The system will allow participants to collaborate on improving models in a decentralized manner without the need for a trusted third-party. We focus on applying DaiMoN for collaboratively improving classification models based on deep learning. However, the presented system can be used with other classes of machine learning models with minimal to no modification.

In traditional blockchains, proof-of-work (PoW)~\cite{nakamoto2008bitcoin} incentivizes people to participate in the consensus protocol for a reward and, as a result, the network becomes more secure as more people participate. In DaiMoN, we introduce the concept of \textit{proof-of-improvement} (PoI). PoI incentivizes people to participate in improving machine learning models for a reward and, as an analogous result, the models on the network become better as more people participate.

One example of a current centralized system that incentivizes data scientists to improve machine learning models for rewards is the Kaggle Competition system~\cite{kaggle}, where a sponsor puts up a reward for contestants to compete to increase the accuracy of their models on a test dataset. The test dataset inputs are given while labels are withheld to prevent competitors from overfitting to the test dataset.

In this example, a sponsor and competitors rely on Kaggle to keep the labels secret. If someone were to hack or compromise Kaggle's servers and gain access to the labels, Kaggle would be forced to cancel the competition. In contrast, because DaiMoN utilizes a DLT, it eliminates this concern to a large degree, as it does not have to rely on a centralized trusted entity.

However, DaiMoN faces a different challenge: in a decentralized ledger, all data are public. As a result, the public would be able to learn about labels in the test dataset if it were to be posted on the ledger. By knowing test labels, peers may intentionally overfit their models, resulting in models which are not generalizable. To solve the problem, we introduce a novel technique, called \textit{Distance Embedding for Labels} (DEL), which can scramble the labels before putting them on the ledger. DEL preserves the error in a predicted label vector inferred by the classifier with respect to the true test label vector of the test dataset, so there is no need to divulge the true labels themselves.

With DEL, we can realize the vision of PoI over a DLT network. That is, any peer verifier can vouch for the accuracy improvement of a submitted model without having access to the true test labels. 
The proof is then included in a block and appended to the ledger for the record. 

The structure of this paper is as follows: after introducing DEL and PoI, we introduce the DaiMoN system that provides incentive for people to participate in improving machine learning models.

The contributions of this paper include:
\begin{enumerate}
    \item A learnable Distance Embedding for Labels (DEL) function specific to the test label vector of the test dataset for the classifier in question, and performance analysis regarding model accuracy estimation and security protection against attacks. To the best of our knowledge, DEL is the first solution which allows peers to verify model quality without knowing the true test labels.
    \item Proof-of-improvement (PoI), including detailed \textsc{Prove} and \textsc{Verify} procedures.
    \item DaiMoN, a decentralized artificial intelligence model network, including an incentive mechanism. DaiMoN is one of the first proof-of-concept end-to-end systems in distributed machine learning based on decentralized ledger technology.
\end{enumerate}

\section{Distance Embedding for Labels (DEL)\label{sec:leph}}

In this section, we describe our proposed Distance Embedding for Labels (DEL), a key technique by which DaiMoN can allow peers to verify the accuracy of a submitted model without knowing the labels of the test dataset. By keeping these labels confidential, the system prevents model-improving peers from overfitting their models intentionally to the test labels.

\subsection{Learning the DEL Function with Multi-Layer Perceptron}
\label{sec:learningDEL}
Suppose that the test dataset for the given $C$-class classification problem consists of $m$ (input, label) test pairs, and each label is an element in $\mcQ = \{c \in \integers \mid 1 \le c \le C\}$, where $\integers$ denotes the set of integers. For example, the FashionMNIST~\cite{xiao2017fashion} classification problem has $C=10$ image classes and $m=10,000$ (input, label) test pairs, where for each pair, the input is a $28\times28$ greyscale image, and the label is an element in $\mcQ = \{1, 2,\dots, 10\}$.

For a given test dataset, we consider the corresponding \textit{test label vector} $\boldx_t \in \mcQ^m$, which is made of all labels in the test dataset. We seek a $\boldx_t$-specific DEL function $f: \boldx \in \mcQ^m \rightarrow \boldy \in \reals^n$, 
where $\reals$ denotes the set of real numbers, which can approximately preserve distance from a predicted label vector $\boldx \in \mcQ^m$ (inferred by a classification model or a classifier we want to evaluate its accuracy) to $\boldx_t$, where $n \ll m$. For example, we may have $n=256$ and $m=10,000$. The \textit{error} of $\boldx$, or the distance from $\boldx$ to $\boldx_t$, is defined as 
\[
\mathrm e(\boldx, \boldx_t) = \frac{1}{m} \sum_{i=1}^m \mathbbm{1}(x_i\neq x_{t_i}),
\]
where $\mathbbm{1}$ is the indicator function, $\boldx = \{x_1,\dots,x_m\}$ and $\boldx_t = \{x_{t_1},\dots,x_{t_m}\}$.

Finding such a distance-preserving embedding function $f$ is generally a challenging mathematical problem. Fortunately, we have observed empirically that we can learn this $\boldx_t$-specific embedding function using a neural network.

More specifically, to learn an $\boldx_t$-specific DEL function $f$, we train a multi-layer perception (MLP) for $f$ as follows. For each randomly selected $\boldx \in \mcQ^m$, we minimize the loss: 
\begin{align*}
    \mcL_{\boldsymbol\theta}(\boldx, \boldx_t) = | \mre(\boldx, \boldx_t) - \mrd(f(\boldx), f(\boldx_t)) |,
\end{align*}
where $\boldsymbol\theta$ is the MLP parameters, and $\mrd(\cdot,\cdot)$ is a modified cosine \textit{distance} function defined as 
\[ 
\mrd(\boldy_1,\boldy_2) = \left\{
\begin{array}{ll}
      1-\frac{\boldy_1\cdot \boldy_2}{\|\boldy_1\|\|\boldy_2\|} & \boldy_1\cdot \boldy_2 \ge 0 \\
     1 & \text{otherwise}.\\
\end{array} 
\right.
\]

The MLP training finds a distance-preserving low-dimensional embedding function $f$ specific to a given $\boldx_t$. The existence of such embedding is guaranteed by the Johnson-Lindenstrauss lemma~\cite{johnson1984extensions, larsen2017optimality}, under a more general setting which does not have the restriction about the embedding being specific to a given vector.

\subsection{Use of DEL Function}

We use the trained DEL function $f$ to evaluate the accuracy or the error of a classification model or a classifier on the given test dataset without needing to know the true test labels. As defined in the preceding section, for a given test dataset, $\boldx_t \in \mcQ^m$ is the true test label vector of the test dataset. Given a classification model or a classifier, $\boldx \in \mcQ^m$ is a predicted label vector consisting of labels inferred by the classifier on all the test inputs of the test dataset. A verifier peer can determine the error of the predicted label vector $\boldx$ without knowing the true test label vector$\boldx_t$, by checking $\mrd(f(\boldx), f(\boldx_t))$ instead of $\mre(\boldx, \boldx_t)$. This is because these two quantities are approximately equal, as assured by the MLP training, which minimizes their absolute difference. If $\mrd(f(\boldx), f(\boldx_t))$ is deemed to be sufficiently lower than that of the previously known model, then a verifier peer may conclude that the model has improved the accuracy of the test dataset. That is, the verifier peer uses $\mrd(f(\boldx), f(\boldx_t))$ as a proxy for $\mre(\boldx, \boldx_t)$.

Note that the DEL function $f$ is $\boldx_t$-specific. For a different test dataset with a different test label vector $\boldx_t$, we will need to train another $f$. For most model benchmarking applications, we expect a stable test dataset; and thus we will not need to retrain $f$ frequently.

\section{Training and Evaluation of DEL}

In this section, we evaluate how well the neural network approach described above can learn a DEL function $f: \mcQ^m \rightarrow \reals^n$ with $m=10,000$ and $n=256$. We consider a simple multi-layer perceptron (MLP) with 1024 hidden units and a rectified linear unit (ReLU). The output of the network is normalized to a unit length. The network is trained using the Adam optimization algorithm~\cite{kingma2014adam}. The dataset used is FashionMNIST~\cite{xiao2017fashion}, which has $C=10$ classes and $m=10,000$ (input, label) test pairs. The true test label vector $\boldx_f$ is thus composed of these 10,000 test labels.

To generate the data to train the function, we perturb the test label vector $\boldx_t$ by using the \textsc{GenerateData} procedure shown.
\begin{figure}[t]
\begin{algorithmic}[1]
\Procedure{GenerateData}{$\boldx_t$}
\State Pick a random number $v$ in $\{1,2,\dots,m\}$
\State Pick a random set $\boldsymbol{\mcK}$ in $\{1,2,\dots,m\}^v$
\State Initialize $\boldx$ as $\{{x_1,x_2,\dots,x_m}\}$ with $\boldx \gets \boldx_t$
\For{$k \in \boldsymbol{\mcK}$}
    \State Pick a random number $c$ in $\{1,2,\dots,C\}$
    \State $x_k \gets c$
\EndFor
\State \textbf{return} $ \boldx$
\EndProcedure
\end{algorithmic}
\label{alg:generatedata}
\end{figure}
First, the procedure picks $v$, the number of labels in $\boldx_t$ to switch out, and generates the set of indices $\boldsymbol{\mcK}$, indicating the positions of the label to replace. It then loops through the set $\boldsymbol{\mcK}$. For each $k \in \boldsymbol{\mcK}$, it generates the new label $c$ to replace the old one. Note that with this procedure, the new label $c$ can be the same as the old label.

\begin{figure}[b]
    \centering
    \includegraphics[width=0.85\linewidth]{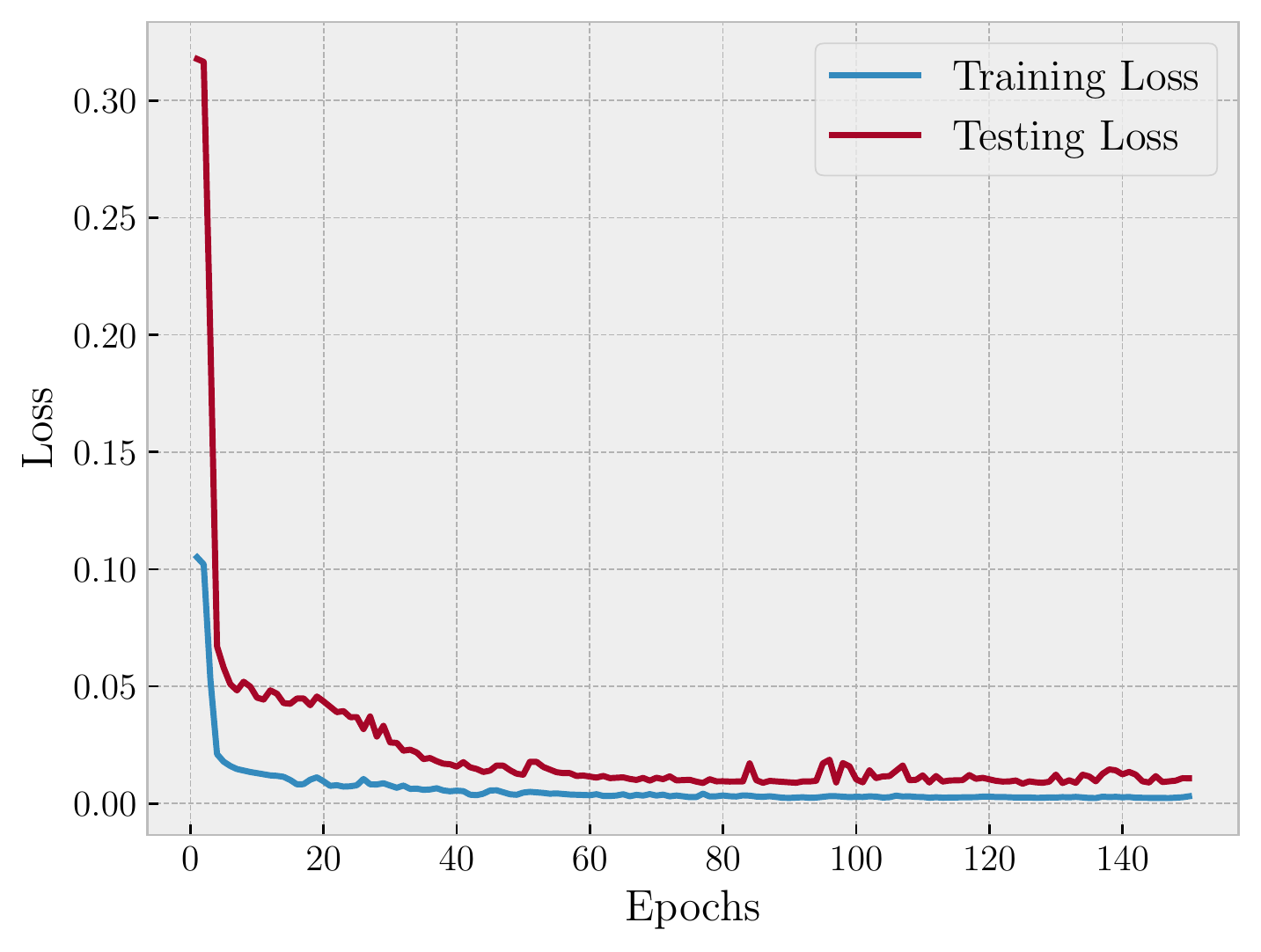}
    \caption{The training and testing loss as the number of epochs increases.}
    \label{fig:errorvsepochlepf}
\end{figure}

We use the procedure to generate the training dataset and the test dataset for the MLP. Figure~\ref{fig:errorvsepochlepf} shows the convergence of the network in learning the function $f$. We see that as the number of epochs increases, both training and testing loss decrease, suggesting that the network is learning the function.

\begin{figure}[h!]
    \centering
    \includegraphics[width=0.85\linewidth]{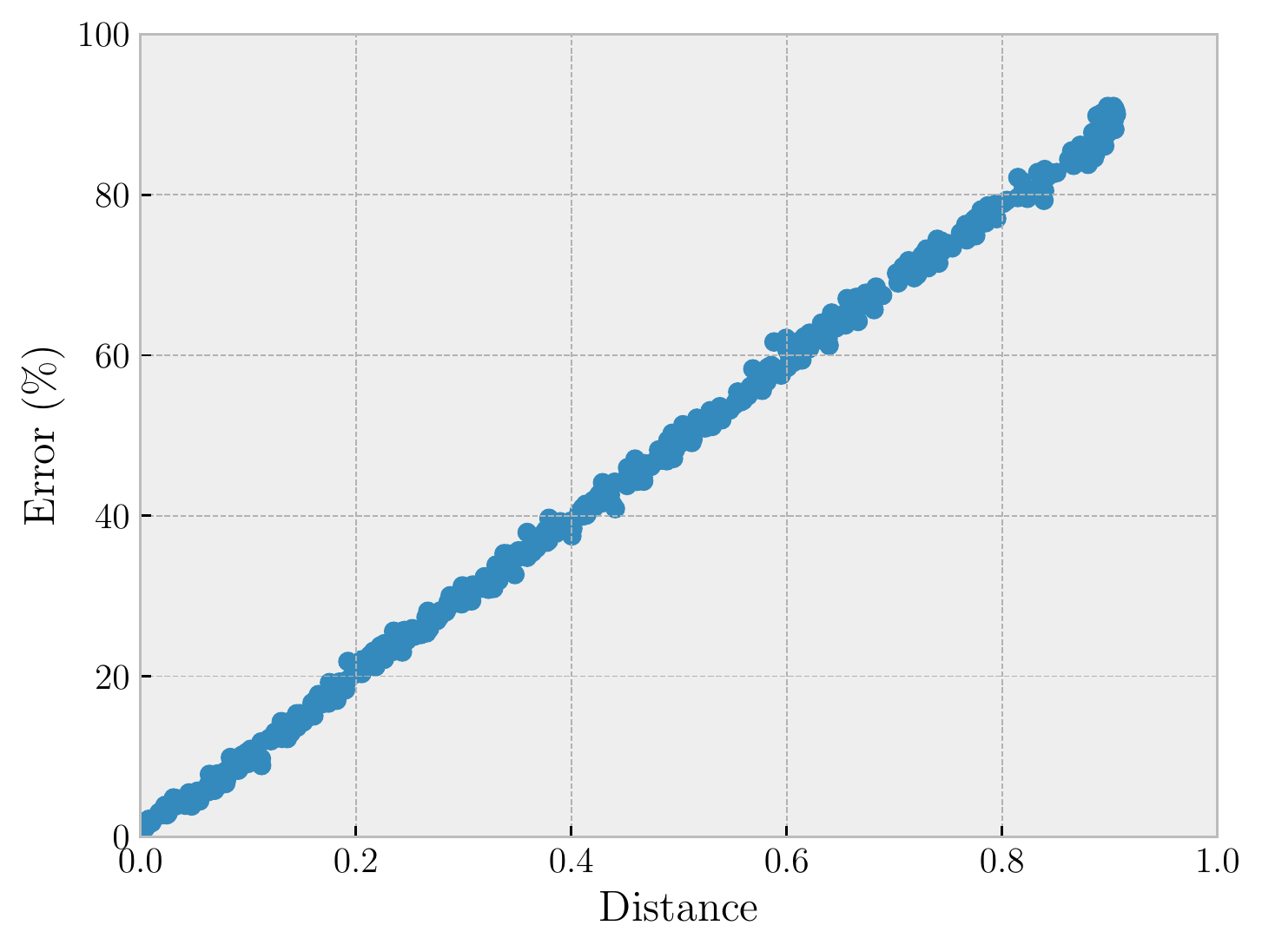}
    \caption{Correlation between error (\%) in $\boldx$ with respect to $\boldx_t$ and the distance between them in the embedding space under $f$.}
    \label{fig:errorvsdist}
\end{figure}


After the neural network has been trained, we evaluate how well the learned $f$ can preserve error in a predicted label vector $\boldx$ inferred by the classifier. Figure~\ref{fig:errorvsdist} shows the correlation between the error and the distance in the embedding space under $f$. We see that both are highly correlated. 



\section{Analysis on Defense Against Brute-force Attacks}

In this section, we show that it is difficult for an attacker to launch a brute-force attack on DEL. To learn about the test label vector $\boldx_t \in \mcQ^m$ that produces $\boldy_t \in \reals^n$ under a known $f: \boldx \in \mcQ^m \rightarrow \boldy \in \reals^n$, the attacker's goal is to find $\boldx$ such that $\mrd(f(\boldx),\boldy_t) < \epsilon$, for a small $\epsilon$. There are $C^m$ possible instances of $\boldx$ to try, where $C$ is the number of classes. Note $C^m$ can be very large. For example, for a test dataset of 10 classes and 10,000 samples, we have $C=10$, $m=10,000$ and $C^m=10^{10000}$. The attacker may use the following brute-force algorithm: 

\begin{figure}[t]
\begin{algorithmic}[1]
\Procedure{BruteForceAttack}{$\boldy_t, \epsilon, q$}
\State Pick a random set $\boldsymbol{\mcX}_0$ of $q$ values in $\mcQ^m$
\For{$\boldx \in \boldsymbol\mcX_0$}
\If{$\mathrm d(f(\boldx),\boldy_t) < \epsilon$}
\State \textbf{return} $ \boldx$
\EndIf
\EndFor
\EndProcedure
\end{algorithmic}
\label{alg:bruteforce}
\end{figure}

The success probability ($\alpha$) of this attack where at least one out of $q$ tried instances for $\boldx$ is within the $\epsilon$ distance of $\boldx_t$ is
$$\alpha = 1-(1-p)^q \approx pq,$$
where $p$ is the probability that $\mathrm d(f(\boldx),\boldy_t) < \epsilon$.

We now derive $p$ and show its value is exceedingly small for a small $\epsilon$, even under moderate values of $n$. Assume that the outputs of $f$ is uniformly distributed on a unit ($n$-1)-sphere or equivalently normally distributed on an $n$-dimension euclidean space~\cite{stam1982limit}. Suppose that $\boldy_t = f(\boldx_t)$. We align the top of the unit ($n$-1)-sphere at $\boldy_t$. Then, $p$ is the probability of a random vector on a ($n$-1)-hemisphere falling onto the cap~\cite{li2011concise} which is 
\begin{align*}
p = \mathrm I_{\sin^2\beta}(\frac{n-1}{2},\frac{1}{2}),
\end{align*}
where $n$ is the dimension of a vector, $\beta$ is the angle between $\boldx_t$ and a vector on the sphere, and $\mathrm I_{x}(a,b)$ is the regularized incomplete beta function defined as: 
\begin{align*}
\mrI_{x}(a,b) & ={\frac {\mathrm {B} (x;\,a,b)}{\mathrm {B} (a,b)}}.
\end{align*}
In the above expression, $\mathrm {B} (x;\,a,b)$ is the incomplete beta function, and $\mathrm {B} (a,b)$ is the beta function defined as:
\begin{align*}
\mathrm {B} (x;\,a,b) & =\int _{0}^{x}t^{a-1}\,(1-t)^{b-1}\,dt, \\
\mathrm {B} (a,b) & =\int _{0}^{1}t^{a-1}\,(1-t)^{b-1}\,dt.
\end{align*}

\begin{figure}[t]
    \centering
    \includegraphics[width=0.85\linewidth]{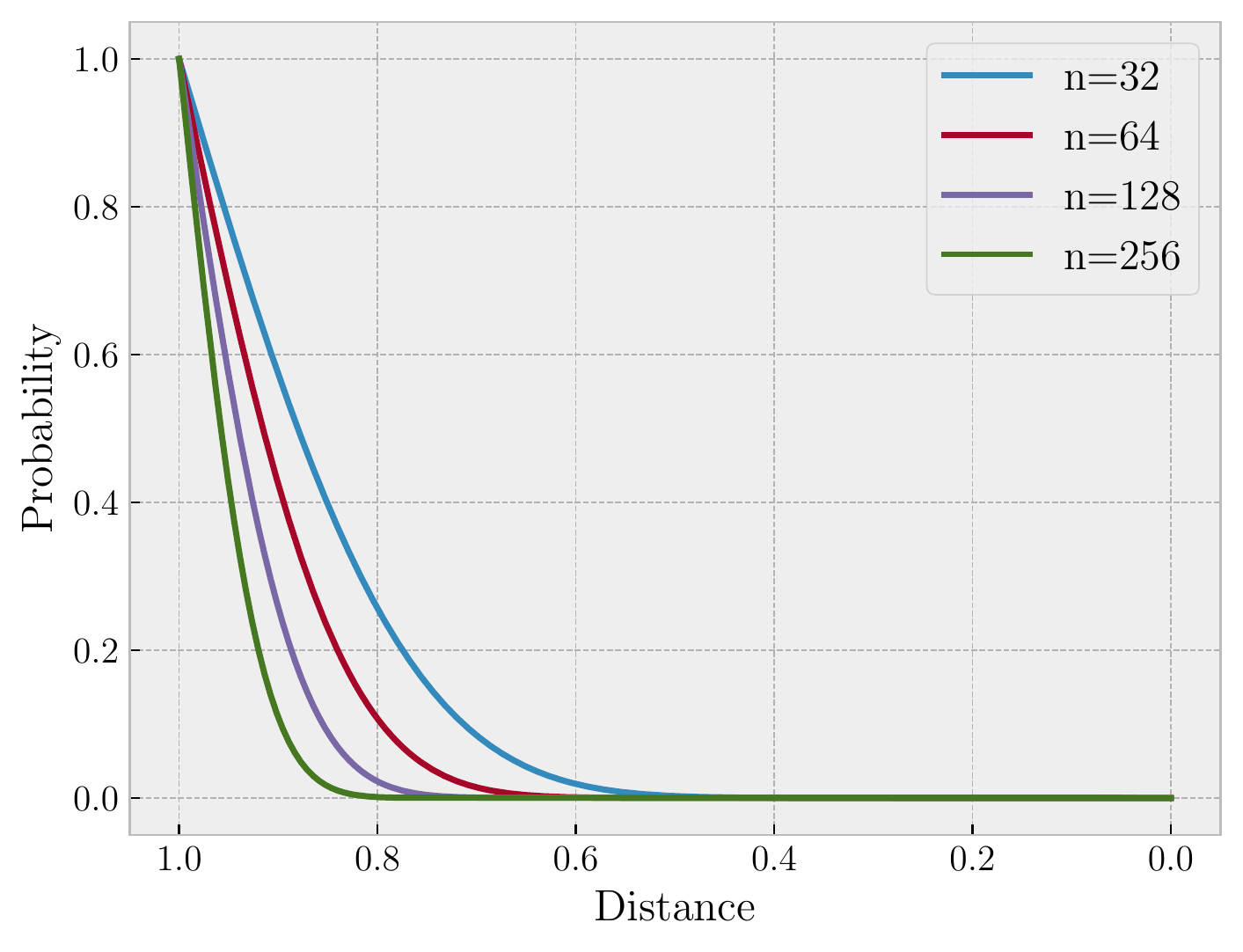}
    \caption{The probability $p$ as distance decreases for varying $n$.}
    \label{fig:pvsangle}
\end{figure}

Figure~\ref{fig:pvsangle} shows the probability $p$ as the distance ($\epsilon$) from $\boldx_t$ decreases for different values of $n$. We observe that for a small $\epsilon$, this probability is exceedingly low and thus to guarantee attacker's success ($\alpha=1$), the number of samples ($q=\alpha/p=1/p$) of $\boldx$ needed to be drawn randomly is very high. For instance, for a 10\% error rate, $\epsilon=0.10$ and $n=32$, the probability $p$ is $6.12\times10^{-13}$ and the number of trials $q$ needed to succeed is $1.63\times10^{12}$. In addition, the higher the $n$, the smaller the $p$ and the larger the $q$. For example, for a 10\% error rate, $\epsilon=0.10$ and $n=256$, the probability $p$ is $3.33\times10^{-93}$ and the number of trials $q$ needed to succeed is $3.01\times10^{92}$.

\section{Analysis on Defense Against Inverse-mapping Attacks}

In this section, we provide an analysis on defense against attacks attempting to recover the original test label vector $\boldx_t$ from $\boldy_t=f(\boldx_t)$. We consider the case that the attacker tries to learn an inverse function $f^{-1}:\boldy \in \reals^n \rightarrow \boldx \in \mcQ^m$ using a neural network. Suppose that the attacker uses a multi-layer perceptron (MLP) for this with 1024 hidden units and a rectified linear unit (ReLU). The network is trained using Adam optimization algorithm~\cite{kingma2014adam}. The loss function used is the squared error function:
\[
\mcL_{\boldsymbol\theta}(\boldy,\boldy_t) = \|f^{-1}(\boldy)-f^{-1}(\boldy_t)\|_2^2.
\]
We generate the dataset using the two procedures: \textsc{GenerateInverseDataNearby} and \textsc{GenerateInverseDataRandom} shown. The former has the knowledge that the test label vector $\boldx_t$ is nearby, and the latter does not. We train the neural network to find the inverse function $f^{-1}$ and compare how the neural network learns from these two generated datasets.

\begin{figure}[t]
\begin{algorithmic}[1]
\Procedure{GenerateInverseDataNearby}{$\boldx_t, f$}
\State $\boldx \gets \textsc{GenerateData}(\boldx_t)$
\State $\boldy \gets f(\boldx)$
\State \textbf{return} $ \{\boldy,\boldx\}$
\EndProcedure
\end{algorithmic}
\label{alg:generateinversedataneartarget}
\end{figure}

The \textsc{GenerateInverseDataNearby} procedure generates a perturbation of the test label vector $\boldx_t$, passes it through the function $f$, and returns a pair of the input $\boldy$ and the target output vector $\boldx$ used to learn the inverse function $f^{-1}$.

\begin{figure}[t]
\begin{algorithmic}[1]
\Procedure{GenerateInverseDataRandom}{$f, C$}
\State Pick a random number $\boldx$ in $\{1,2,\dots,C\}^m$
\State $\boldy \gets f(\boldx)$
\State \textbf{return} $ \{\boldy,\boldx\}$
\EndProcedure
\end{algorithmic}
\label{alg:generateinversedata}
\end{figure}

The \textsc{GenerateInverseDataRandom} procedure generates a random label vector $\boldx_t$ where each element of the vector has a value representing one of the C classes, passes it through the function $f$ and returns a pair of the input $\boldy$ and the target output vector $\boldx$ used to learn the inverse function $f^{-1}$.

\begin{figure}[t]
    \centering
    \includegraphics[width=0.85\linewidth]{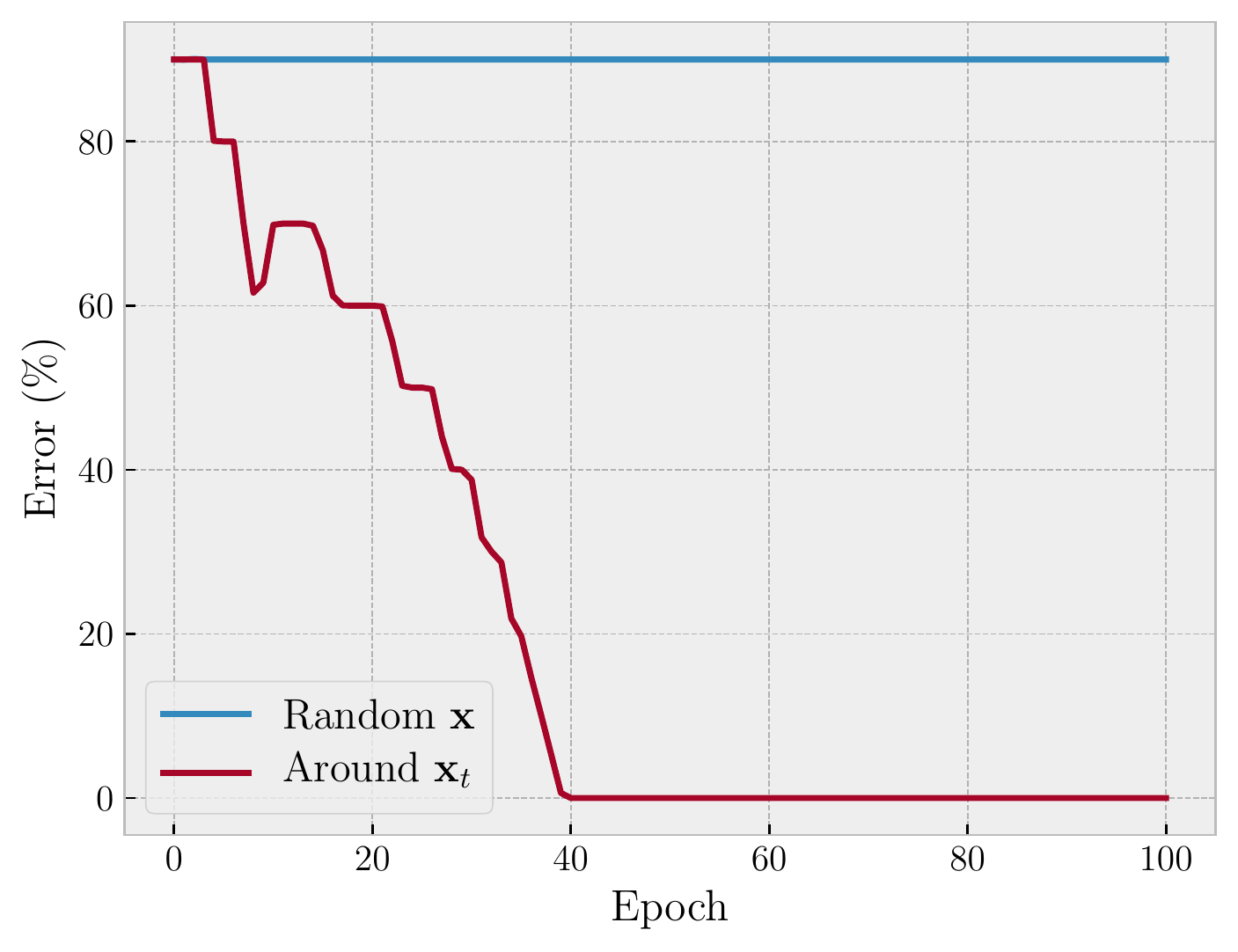}
    \caption{Error ($\mre(f^{-1}(\boldy_t),\boldx_t)$) in percentage as the number of epochs increases for data generated with random $\boldx$ and near $\boldx_t$.}
    \label{fig:errorvsepoch}
\end{figure}
Figure~\ref{fig:errorvsepoch} shows the error $\mre(f^{-1}(\boldy_t),\boldx_t)$ as the number of training epochs increases. On one hand, using the data generated without the knowledge of the test label vector $\boldx_t$ using the \textsc{GenerateInverseDataRandom} procedure, we see that the network does not reduce the error as it trains. This means that it does not succeed in learning the inverse function $f^{-1}$ and therefore, it will not be able to recover the test label $\boldx_t$ from its output vector $\boldy_t$. On the other hand, using the data generated with the knowledge of the test label vector $\boldx_t$ using the \textsc{GenerateInverseNearby} procedure, we see that the network does reduce the the error as it trains and has found the test label vector $\boldx_t$ from its output vector $\boldy_t$ at around 40 epochs. This experiment gives an empirical evidence that without the knowledge of $\boldx_t$, it is hard to find $f^{-1}$.

\section{Proof-of-Improvement}

In this section, we introduce the concept of \textit{proof-of-improvement} (PoI), a key mechanism supporting DaiMoN. PoI allows a prover $\mcP$ to convince a verifier $\mcV$ that a model $\mrM$ improves the accuracy or reduces the error on the test dataset via the use of DEL without the knowledge of the true test label vector $\boldx_t$. PoI is characterized by the \textsc{Prove} and \textsc{Verify} procedures shown.

As a part of the system setup, a prover $\mcP$ has a public and private key pair ($\mathrm{pk}_{\mcP}$, $\mathrm{sk}_{\mcP}$) and a verifier $\mcV$ has a public and private key pair ($\mathrm{pk}_{\mcV}$, $\mathrm{sk}_{\mcV}$). Both are given our learnt DEL function $f(\cdot)$, $\boldy_t=f(\boldx_t)$, the set of $m$ test inputs $\boldZ = \{\boldz\}_{i=1}^m$, and the current best distance $d_c$ achieved by submitted models, according to the distance function $\mrd(\cdot, \cdot)$ described in Section~\ref{sec:leph}.

Let $\mathrm{digest}(\cdot)$ be the message digest function such as IPFS hash~\cite{benet2014ipfs}, MD5~\cite{rivest1992md5}, SHA~\cite{bertoni2013keccak} and CRC~\cite{peterson1961cyclic}. 
Let $\{\cdot\}_{\mathrm{sk}}$ denotes a message signed by a secret key $\mathrm{sk}$.

Let $\mrM$ be the classification model for which $\mcP$ will generate a PoI proof $\pi_\mcP$. The model $\mrM$ takes an input and returns the corresponding predicted class label. The \textsc{Prove} procedure called by a prover $\mcP$
generates the digest of $\mrM$ and calculates the DEL function output of the predicted labels of the test dataset $\boldZ$ by $\mrM$. The results are concatenated to form the body of the proof, which is then signed using the prover's secret key $\mathrm{sk}_{\mcP}$. The PoI proof $\pi_\mcP$ shows that the prover $\mcP$ has found a model $\mrM$ that could reduce the error on a test dataset.
\begin{figure}[t]
\begin{algorithmic}[1]
\Procedure{Prove}{$\mrM$}{ $\rightarrow \pi_\mcP$}
\State $\boldg \gets \mathrm{digest}(\mrM)$
\State $\boldy \gets f(\mrM(\boldZ))$
\State \textbf{return} $ \{\boldg,\boldy,\mathrm{pk}_\mcP\}_{\mathrm{sk}_{\mcP}}$
\EndProcedure
\end{algorithmic}
\label{alg:prove}
\end{figure}

To verify, the verifier $\mcV$ runs the following procedure to generate the verification proof $\pi_\mcV$, the proof that the verifier $\mcV$ has verified the PoI proof $\pi_\mcP$ generated by the prover $\mcP$.
\begin{figure}[t]
\begin{algorithmic}[1]
\Procedure{Verify}{$\mrM, \pi_\mcP, d_\text{c}, \delta$}{ $\rightarrow \pi_\mcV$}
\State Verify the signature of $\pi_\mcP$ with $\pi_\mcP.\mathrm{pk}_{\mcP}$
\State Verify the digest: $\pi_\mcP.\boldg = \mathrm{digest}(\mrM)$
\State Verify the DEL function output: $\pi_\mcP.\boldy = f(\mrM(\boldZ))$
\State Verify the distance: $\mrd(\pi_\mcP.\boldy, \boldy_t) < d_\text{c} - \delta, \delta \ge 0$
\If{all verified}
    \State \textbf{return} $\{\pi_\mcP,d_c,\delta,\mathrm{pk}_\mcV\}_{\mathrm{sk}_{\mcV}}$
\EndIf
\EndProcedure
\end{algorithmic}
\label{alg:verify}
\end{figure}
The procedure first verifies the signature of the proof with public key $\pi_\mcP.\mathrm{pk}_{\mcP}$ of the prover $\mcP$. Second, it verifies that the digest is correct by computing $\mathrm{digest}(\mrM)$ and comparing it with the digest in the proof $\pi_\mcP.\boldg$. Third, it verifies the DEL function output by computing $f(\mrM(\boldZ))$ and comparing it with the the DEL function output in the proof $\pi_\mcP.\boldy$. Lastly, it verifies the distance by computing $\mrd(\pi_\mcP.\boldy, \boldy_t)$ and sees if it is lower than the current best with a margin of $\delta \ge 0$, where $\delta$ is an improvement margin commonly agreed upon among peers. If all are verified, the verifier generates the body of the verification proof by concatenating the PoI proof $\pi_\mcP$ with the current best distance $d_c$ and $\delta$. Then, the body is signed with the verifier's secret key $\mathrm{sk}_{\mcV}$, and the verification proof is returned.

\section{The DaiMoN System}

In this section, we describe the DaiMoN system that incentivizes participants to improve the accuracy of models solving a particular problem.
In DaiMoN, each classification problem has its own DaiMoN blockchain with its own token. An append-only ledger maintains the log of improvements for that particular problem. A problem defines inputs and outputs which machine learning models will solve for. We call this the problem definition. For example, a classification problem on the FashionMNIST dataset~\cite{xiao2017fashion} may define an input $\boldz$ as a 1-channel $1\times28\times28$ pixel input whose values are ranging from 0 to 1, and an output $\boldx$ as 10-class label ranging from 1 to 10:
\begin{align*}
& \{\boldz\in \reals^{1\times28\times28} \mid 0\le \boldz \le 1\},\\
& \{\boldx\in \integers \mid 1\le \boldx \le 10\}.
\end{align*}

Each problem is characterized by a set of test dataset tuples. Each tuple ($\boldZ,f,\boldy_t$) consists of the test inputs $\boldZ = \{\boldz\}_{i=1}^m$, the DEL function $f$, and the DEL function output $\boldy_t = f(\boldx_t)$ on the true test label vector $\boldx_t$.

A participant is identified by its public key $\mathrm{pk}$ with the associated private key $\mathrm{sk}$.
There are six different roles in DaiMoN: problem contributors, model improvers, validators, block committers, model runners, and model users. A participant can be one or more of these roles. We now detail each role below:

\textbf{Problem contributors} contribute test dataset tuples to a problem. They can create a problem by submitting a problem definition and the first test dataset tuple. See Section~\ref{sec:consensus} on how additional test tuples can be added.

\textbf{Model improvers} compete to improve the accuracy of the model according to the problem definition defined in the chain. A model improver generates a PoI proof for the improved model and submit it.

\textbf{Validators} validate PoI proofs, generate a verification proof and submit it as a vote on the PoI proofs. Beyond being a verifier for verifying PoI, a validator submits the proof as a vote.

\textbf{Block committers} create a block from the highest voted PoI proof and its associated verification proofs and commit the block.

\textbf{Model runners} run the inference on the latest model given inputs and return outputs and get paid in tokens.

\textbf{Model users} request an inference computation from model runners with an input and pay for the computation in tokens.

\subsection{The Chain}

Each chain consists of two types of blocks: Problem blocks and Improvement blocks. 
\textbf{A Problem block} contains information, including but not limited to: the block number, the hash of the parent block, the problem definition, the test dataset tuples, and the block hash.
\textbf{An Improvement block} contains information, including but not limited to: the block number, the hash of the parent block, PoI proof $\pi_\mcP$ from model improver $\mcP$, verification proofs $\{\pi_\mcP\}$ from validators $\{\mcV\}$, and the block hash. The chain must start with a Problem block that defines the problem, followed by Improvement blocks that record the improvements made for the problem.


\subsection{The Consensus\label{sec:consensus}}

After a DaiMoN blockchain is created, there is a problem definition period $T_p$. In this period, any participant is allowed to add test dataset tuples into the mix. After a time period $T_p$ has passed, a block committer commits a Problem block containing all test dataset tuples submitted within the period to the chain. 

After the Problem block is committed, a competition period $T_b$ begins. During this period, a model improver can submit the PoI proof of his/her model. A validator then validates the PoI proof and submit a verification proof as a vote. For each PoI proof, its associated number of unique verification proofs are tracked. At the end of each competition period, a block committer commits an Improvement block containing the model with the highest number of unique verification proofs, and the next competition period begins.






\subsection{The Reward}
\begin{figure}[t]
    \centering
    \includegraphics[width=0.85\linewidth]{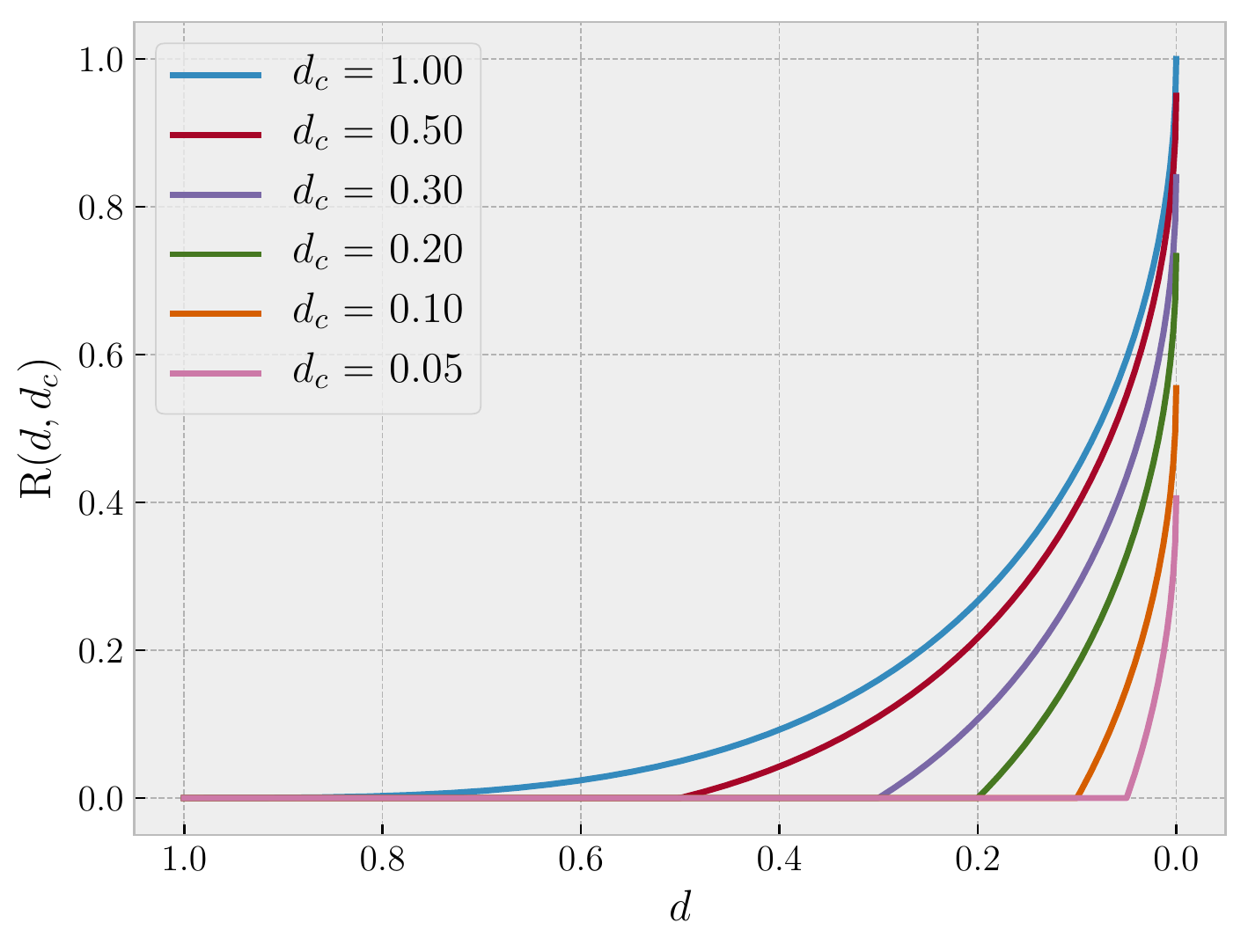}
    \caption{The reward function $\mrR(d,d_c)$ as the distance $d$ decreases for varying current best distance $d_c$ for $a=3$.}
    \label{fig:reward}
\end{figure}

Each committed block rewards tokens to the model improver and validators. The following reward function, or similar ones, can be used:
\[
\mrR(d,d_c) = \mrI_{1-d}(a,\frac{1}{2}) - \mrI_{1-d_c}(a,\frac{1}{2}),
\]
where $\mrI_{\cdot}(\cdot,\cdot)$ is the regularized incomplete beta function, $d$ is the distance of the block, $d_c$ is the current best distance so far, and $a$ is a parameter to allow for the adjustment to the shape of the reward function. Figure~\ref{fig:reward} shows the reward function as the distance $d$ decreases for different current best distance $d_c$ for $a=3$. We see that more and more tokens are rewarded as the distance $d$ reaches 0, and the improvement gap $d_c - d$ increases.

Each validator is given a position as it submits the validation proof: the $s$-th validator to submit the validation proof is given the $s$-th position. The validator's reward is the model improver's reward scaled by $2^{-s}$: $\mrR(d,d_c) 2^{-s}$,
where $s \in \integers_{>0}$ is the validator's position, and $\integers_{>0}$ denotes the set of integers greater than zero.
This factor encourages validators to compete to be the first one to submit the validation proof for the PoI proof in order to maximize the reward. Two is used as a base of the scaling factor here since $\sum_{s=1}^\infty 2^{-s} = 1.$


\subsection{The Market}

In order to increase the value of the token of each problem, there should be demand for the token. One way to generate demand for the token is to allow it to be used as a payment for inference computation based on the latest model committed to the chain. To this end, model runners host the inference computation. Each inference call requested by users is paid for by the token of the problem chain that the model solves. Model runners automatically upgrade the model, as better ones are committed to the chain. The price of each call is set by the market according to the demand and supply of each service. This essentially determines the value of the token, which can later be exchanged with other cryptocurrencies or tokens on the exchanges. As the demand for the service increases, so will the token value of the problem chain.

Model runners periodically publish their latest services containing the price for the inference computation of a particular model. Once a service is selected, model users send a request with the payment according to the price specified. Model runners then verify the request from the user, run the computation, and return the result.

To keep a healthy ecosystem among peers, a reputation system may be used to recognize good model runners and users, and reprimand bad model runners and users. Participants in the network can upvote good model runners and users and downvote bad model runners and users. 

\subsection{System Implementation}

DaiMoN is implemented on top of the Ethereum blockchain~\cite{wood2014ethereum}. In this way, we can utilize the security and decentralization of the main Ethereum network. The ERC-20~\cite{vogelsteller2018erc} token standard is used to create a token for each problem chain. Tokens are used as an incentive mechanism and can be exchanged. Smart contracts are used to manage the DaiMoN blockchain for each problem. 

Identity of a participant is represented by its Ethereum address. Every account on Ethereum is defined by a pair of keys, a private key and public key. Accounts are indexed by their address, which is the last 20 bytes of the Keccak~\cite{bertoni2013keccak} hash of the public key.


The position of the verifiers is recorded and verified on the Ethereum blockchain. As a verifier submits a vote on the smart contract on the Ethereum blockchain, his/her position is recorded and used to calculate the reward for verifier.


The InterPlanetary File System (IPFS)~\cite{benet2014ipfs} is used to store and share data files that are too big to store on the Ethereum blockchain. Files such as test input files and model files are stored on IPFS and only their associated IPFS hashes are stored in the smart contracts. Those IPFS hashes are then used by participants to refer to the files and download them. Note that since storing files on IPFS makes it public, it is possible that an attacker can find and submit the model before the creator of the model. To prevent this, model improvers must calculate the IPFS hash of the model and register it with the smart contract on the Ethereum blockchain before making the model available on IPFS.


\section{Discussion}

One may compare a DEL function to an encoder of an autoencoder~\cite{rumelhart1985learning}. An autoencoder consists of an encoder and a decoder. The encoder maps an input to a lower-dimensional embedding which is then used by the decoder to reconstruct the original input. Although a DEL function also reduces the dimensionality of the input label vector, it does not require the embedding to reconstruct the original input and it adds the constraint that the output of the function should preserve the error or the distance of the input label vector to a specific test label vector $\boldx_t$. In fact, for our purpose of hiding the test labels, we do not want the embedding to reconstruct the original input test labels. Adding the constraint to prevent the reconstruction may help further defense against the inverse-mapping attacks and can be explored in future work.

Note that a model with closer distance to the test label vector ($\boldx_t$) in the embedding space may not have better accuracy. This results in a reward being given to a model with worse accuracy than the previous best model. This issue can be mitigated by increasing the margin $\delta$. With the appropriate $\delta$ setting, this discrepancy should be minimal. Note also that as the model gets better, it will be easier for an attacker to recover the true test label vector ($\boldx_t$). To mitigate this issue, multiple DEL and reward functions may be used at various distance intervals.


By building DaiMoN on top of Ethereum, we inherit the security and decentralization of the main Ethereum network as well as the limitations thereof. We now discuss the security of each individual DaiMoN blockchain. An attack to consider is the Sybil attack on the chain, in which an attacker tries to create multiple identities (accounts) and submit multiple verification proofs on an invalid PoI proof. Since each problem chain is managed using Ethereum smart contracts, there is an inherent gas cost associated with every block submission. Therefore, it may be costly for an attacker to overrun the votes of other validators. The more number of validators for that chain, the higher the cost is. In addition, this can be thwarted by increasing the cost of each submission by requiring validators to also pay Ether as they make the submission. All in all, if the public detects signs of such behavior, they can abandon the chain altogether. If there is not enough demand in the token, the value of the tokens will depreciate and the attacker will have less incentives to attack.

Since we use IPFS in the implementation, we are also limited by the limitations of IPFS: files stored on IPFS are not guaranteed to be persistent. In this case, problem contributors and model improvers need to make sure that their test input files and model files are available to be downloaded on IPFS. In addition to IPFS, other decentralized file storage systems that support persistent storage at a cost such as Filecoin~\cite{benet2017filecoin}, Storj~\cite{wilkinson2014storj}, etc. can be used.

\section{Related Works}

One area of related work is on data-independent locality sensitive hashing (LSH)~\cite{indyk1998approximate} and data-dependent locality preserving hashing (LPH)~\cite{indyk1997locality, zhao2014locality}. LSH hashes input vectors so that similar vectors have the same hash value with high probability. There are many algorithms in the family of LSH. One of the most common LSH methods is the random projection method called SimHash~\cite{charikar2002similarity}, which uses a random hyperplane to hash input vectors.

Locality preserving hashing (LPH) hashes input vectors so that the relative distance between the input vectors is preserved in the relative distance between of the output vectors; input vectors that are closer to each other will produce output vectors that are closer to each other in the output space. The DEL function presented in this paper is in the family of LPH functions. While most of the work on LSH and LPH focuses on dimensionality reduction for nearest neighbor searches, DEL is novel that it focuses on learning an embedding function that preserves the distance to the test label vector of the test dataset. For the purpose of hiding the test label vector from verifier peers, it is appropriate that DEL's distance preserving is specific to this test vector. By being specific, finding the DEL function becomes easier. 

Another area of related work is on blockchain and AI. There are numerous projects covering this area in recent years. These include projects such as SingularityNET~\cite{goertzel2017singularitynet}, Effect.ai~\cite{jesse2018effect} and Numerai~\cite{craib2017numeraire}. SingularityNET and Effect.ai are decentralized AI marketplace platforms where anyone can provide AI services for use by the network, and receive network tokens in exchange. This is related to the DaiMoN market, where model runners run inference computation for model users in exchange for tokens. Numerai has an auction mechanism where participants stake network tokens to express confidence in their models' performance on forthcoming new data. The mechanism will reward participants according to the performance of their models on unspecified yet future test data, their confidence, and their stake. While these platforms all use a single token, DaiMoN has a separate token for each problem. It also introduces the novel ideas of DEL and PoI, allowing the network to fairly reward participants that can prove that they have a model that improves the accuracy for a given problem.

\section{Conclusion}

We have introduced DaiMoN, a decentralized artificial intelligence model network. DaiMoN uses a Distance Embedding for Labels (DEL) function. DEL embeds the predicted label vector inferred by a classifier in a low-dimensional space where its error or its distance to the true test label vector of the test dataset is approximately preserved. Under the embedding, DEL hides test labels from peers while allowing them to assess the accuracy improvement that a model makes. We present how to learn DEL, evaluate its effectiveness, and present the analysis of DEL's resilience against attacks. This analysis shows that it is hard to launch a brute-force attack or an inverse-mapping attack on DEL without knowing a priori a good estimate on the location of the test label vector, and that the hardness can be increased rapidly by increasing the dimension of the embedding space.

DEL enables \textit{proof-of-improvement} (PoI), the core of DaiMoN. Participants use PoI to prove that they have found a model that improves the accuracy of a particular problem. This allows the network to keep an append-only log of model improvements and reward the participants accordingly. DaiMoN uses a reward function that scales according to the increase in accuracy a new model has achieved on a particular problem. We hope that DaiMoN will spur distributed collaboration in improving machine learning models.

\section{Acknowledgment}
This work is supported in part by the Air Force Research Laboratory under agreement number FA8750-18-1-0112 and a gift from MediaTek USA.

\small
\bibliography{paper}
\bibliographystyle{IEEEtran}

\end{document}